%% file: acl_latex.tex
\documentclass[11pt]{article}

\usepackage[final]{acl}

\usepackage{times}
\usepackage{latexsym}

\usepackage[T1]{fontenc}

\usepackage[utf8]{inputenc}

\usepackage{microtype}

\usepackage{inconsolata}

\usepackage{graphicx}

\usepackage{bm}
\usepackage{amsmath}
\usepackage{booktabs}    
\usepackage{subcaption}  
\usepackage{wrapfig}
\usepackage{hyperref}
\usepackage{url}
\usepackage{amsmath}
\usepackage{booktabs}
\usepackage{amsfonts}
\usepackage{amssymb}
\usepackage{threeparttable}
\usepackage{booktabs}
\usepackage{multirow}
\usepackage{subcaption}
\usepackage{adjustbox}

\usepackage{algorithm}
\usepackage{algpseudocode}
\usepackage{enumitem}
\usepackage{newfloat}
\usepackage{listings}
\usepackage{subfiles} 

\newcommand{\kv}{OjaKV}

\usepackage[framemethod=default]{mdframed}
\usepackage{xfp} 
\newcommand{\avgXI}[1]{\fpeval{round((#1)/11,2)}}

\usepackage{color}

\newmdenv[
  linecolor=black,
  linewidth=0.5pt,
  roundcorner=2pt,
  backgroundcolor=gray!5, 
  topline=false,
  bottomline=false,
  rightline=false,
  leftmargin=5pt,
  rightmargin=5pt,
  innerleftmargin=10pt,
  innerrightmargin=10pt,
  innertopmargin=10pt,
  innerbottommargin=10pt,
  frametitlefont=\bfseries,
]{casestudybox}

\title{OjaKV: Context-Aware Online Low-Rank KV Cache Compression}

\author{
  Yuxuan Zhu\textsuperscript{1}\thanks{Equal contribution.}, David H. Yang\textsuperscript{1}\textsuperscript{*},  Mohammad Mohammadi Amiri\textsuperscript{1}, \\ 
  \textbf{Keerthiram Murugesan\textsuperscript{2},  Tejaswini Pedapati\textsuperscript{2} 
  , Pin-Yu Chen\textsuperscript{2}} \\
  \textsuperscript{1}Rensselaer Polytechnic Institute \quad \textsuperscript{2}IBM Research\\
  {\tt \{yangd13, zhuy27, mamiri\}@rpi.edu} \quad {\tt tejaswinip@us.ibm.com} \\
  {\tt \{keerthiram.murugesan, pin-yu.chen\}@ibm.com}
}

\begin{document}
\maketitle
\begin{abstract}
The expanding long-context capabilities of large language models are constrained by a significant memory bottleneck: the key-value (KV) cache required for autoregressive generation.\ This bottleneck is substantial; for instance, a Llama-3.1-8B model processing a 32K-token prompt at a batch size of 4 requires approximately 16\,GB for its KV cache, a size exceeding the model's weights.\ While KV-cache compression via low-rank projection is a promising direction, existing methods' rely on a static, offline-learned subspace that performs poorly under data distribution shifts.\ To overcome these limitations, we introduce \textbf{OjaKV}, a novel framework that integrates a strategic hybrid storage policy with online subspace adaptation.\ First, OjaKV recognizes that not all tokens are equally important for compression; it preserves the crucial tokens in full-rank, maintaining high-fidelity anchors for attention.\ Second, for the majority of intermediate tokens, it applies low-rank compression by incrementally adapting the projection basis using Oja’s algorithm for online principal component analysis.\ This adaptation involves a comprehensive update during prefilling and lightweight periodic updates during decoding, ensuring the subspace remains aligned with the evolving context.\ Crucially, our framework is fully compatible with modern attention modules like \textit{FlashAttention}.\ Experiments show that OjaKV maintains or even improves accuracy at high compression ratios.\ In particular, OjaKV achieves its strongest gains on very long-context benchmarks that require complex reasoning, highlighting the importance of online subspace adaptation in dynamically tracking context shifts.\ Furthermore, our approach is compatible with token-selection methods, enabling compounded memory savings.\ These results establish our hybrid framework as a practical, plug-and-play solution for memory-efficient long-context inference without requiring model fine-tuning.\ Code available at \url{https://github.com/zzbright1998/OjaKV}
\end{abstract}

\subfile{Sections/Chapter_1}

\subfile{Sections/Chapter_2}

\subfile{Sections/Chapter_3}

\subfile{Sections/Chapter_4}

\subfile{Sections/Chapter_5}
\subfile{Sections/Chapter_6}

\section*{Limitations}
Our method has several limitations. First, OjaKV introduces fixed hyperparameters (learning rates, buffer size, top-$k$) that may require tuning for different models or tasks. Second, while the Oja update is lightweight, it still incurs additional computational overhead compared to static compression methods. Third, our hybrid selection strategy relies on reconstruction error, which may not perfectly capture token importance for all downstream tasks.

\bibliography{acl}

\appendix

\section{Appendix}

\subsection{Oja's Updating Algorithm}
\label{subsec: algo}
We consolidate the complete OjaKV online updating process into a formal procedure in Algorithm~\ref{alg:online-eigen-attn}. The algorithm outlines the two primary stages of our method: a comprehensive update during the prefill phase and lightweight, periodic updates during the decoding phase.

The \textbf{Prefilling Phase} is designed to adapt the initial, general-purpose projection matrices ($U_{k}$ and $U_v$) to the specific content of the input prompt. First, average pooling with size $p$ is applied to reduce the number of vectors for efficient computation (Line 2). The pooled key and value vectors are then used to perform a batch update on the projection matrices via Oja's rule (Lines 3-4), after which the matrices are re-orthonormalized to maintain their geometric properties (Line 5). Finally, our hybrid storage policy is enacted through an error-weighted selection mechanism: we compute the reconstruction error $\bm{K} - \hat{\bm{K}}$ weighted by query attention from a local window, normalize the scores, and select the top-$k$ tokens with the highest error to be stored at full rank (Lines 6-7).

The \textbf{Decoding Phase} handles the continuous adaptation of the subspace as new tokens are autoregressively generated. At each step, the newly generated key-value pair is temporarily stored in a buffer (Line 11). To maintain efficiency, updates are performed periodically. Every $T$ steps, average pooling is applied to the buffered vectors, followed by another batch Oja update with a more conservative learning rate to ensure stable learning (Lines 13-14). The bases are again re-orthonormalized, and the buffers are cleared for the next cycle (Lines 15-16). This two-phase approach allows OjaKV to make a strong initial adaptation to the context while continuously refining the subspace with minimal overhead during generation.

\begin{algorithm}[!htb]
\caption{\kv{}}
\label{alg:online-eigen-attn}
\begin{algorithmic}[1]
\Require Low rank projection matrices $\bm{U}_{k},\bm{U}_v$; learning rates $\eta$; update buffer size $T$; pool size $p$; top-$k$ for selection
\State \textbf{Prefilling Phase:}
\State Form matrices $\bm{K}, \bm{V}$; apply average pooling with size $p$
\State $\tilde{\bm{K}}\gets \bm{U}_{k}^{\mathsf T}\bm{K}$;\;\;$\bm{U}_{k}\gets \bm{U}_{k}+\eta\bigl(\bm{K}-\bm{U}_{k}{\tilde{\bm{K}}}\bigr)\tilde{\bm{K}}^{\mathsf T}$
\State $\tilde{\bm{V}}\gets \bm{U}_{v}^{\mathsf T}\bm{V}$;\;\;$\bm{U}_{v}\gets \bm{U}_{v}+\eta\bigl(\bm{V}-\bm{U}_{v}\tilde{\bm{V}}\bigr)\bm{\tilde{V}}^{\mathsf T}$
\State $(\bm{U}_{k},\bm{U}_v)\gets\text{Orthonormalise}(\bm{U}_{k},\bm{U}_v)$
\State $\hat{\bm{K}} \gets \bm{U}_{k}\bm{U}_{k}^{\mathsf T}\bm{K}$; \; $\bm{E} \gets |\bm{Q}_{\text{window}}^{\mathsf T}(\bm{K} - \hat{\bm{K}})|$ \Comment{Error-weighted selection}
\State $\mathcal{I}_{\text{full}} \gets \text{Top-}k(\text{Normalize}(\bm{E}))$
\State
\State \textbf{Decoding Phase:}
\For{step $t=1,2,\dots$}
    \State Generate new $(k_t,v_t)$ and append to buffers $\mathcal{B}_k,\mathcal{B}_v$
    \If{$t\bmod T=0$}
        
        \State $\bm{\tilde{K}}\gets \bm{U_{k}}^{\mathsf T}\bm{K}$;\;\;$\bm{U_{k}}\gets \bm{U_{k}}+\eta\bigl(\bm{K}-\bm{U_{k}}\bm{\tilde{K}}\bigr)\bm{\tilde{K}}^{\mathsf T}$
        \State $\bm{\tilde{V}}\gets \bm{U_{v}}^{\mathsf T}\bm{V}$;\;\;$\bm{U}_{v}\gets \bm{U}_{v}+\eta\bigl(\bm{V}-\bm{U}_{v}\tilde{\bm{V}}\bigr)\tilde{\bm{V}}^{\mathsf T}$
        \State $(\bm{U}_{k},\bm{U}_v)\gets\text{Ortho}(\bm{U}_{k},\bm{U}_v)$
        \State Reset $\mathcal{B}_k,\mathcal{B}_v$
    \EndIf
\EndFor
\end{algorithmic}
\end{algorithm}

\subsection{From subspace error to attention and generation error}
\label{sec:theory_guarantee}

We provide a simple end-to-end argument connecting low-rank approximation error, attention perturbation, and downstream generation quality.

Let $k_t \in \mathbb{R}^{d_h}$ be a full-rank key and let $\hat{k}_t$ denote its compressed-and-reconstructed version. Write
\[
\hat{k}_t = k_t + e_t,
\]
where $e_t$ is the reconstruction residual. For a query $q$, the original and perturbed attention logits are
\[
z_t = \frac{q^\top k_t}{\sqrt{d_h}},
\qquad
\hat{z}_t = \frac{q^\top \hat{k}_t}{\sqrt{d_h}}.
\]
Their difference is
\[
|\hat{z}_t - z_t|
=
\frac{|q^\top e_t|}{\sqrt{d_h}}
\le
\frac{\|q\|_2 \|e_t\|_2}{\sqrt{d_h}}
\le
\frac{Q\|e_t\|_2}{\sqrt{d_h}},
\]
where we assume $\|q\|_2 \le Q$.

Now define $z,\hat{z} \in \mathbb{R}^n$ as the full logit vectors over all attended tokens, and let
\[
\alpha = \mathrm{softmax}(z),
\qquad
\hat{\alpha} = \mathrm{softmax}(\hat{z}).
\]
Using standard softmax stability,
\[
\|\hat{\alpha} - \alpha\|_1
\le
2\|\hat{z} - z\|_\infty.
\]
If we let
\[
E = \max_t \|e_t\|_2
\]
over compressed tokens, then
\[
\|\hat{z} - z\|_\infty \le \frac{Q E}{\sqrt{d_h}},
\qquad
\|\hat{\alpha} - \alpha\|_1
\le
\frac{2QE}{\sqrt{d_h}}.
\]

Next, let the attention outputs be
\[
o = \sum_{t=1}^n \alpha_t v_t,
\qquad
\hat{o} = \sum_{t=1}^n \hat{\alpha}_t v_t,
\]
and assume $\|v_t\|_2 \le V$ for all $t$. Then
\[
\begin{aligned}
\|\hat{o} - o\|_2
&=
\left\|
\sum_{t=1}^n (\hat{\alpha}_t - \alpha_t)v_t
\right\|_2 \\
&\le
\sum_{t=1}^n |\hat{\alpha}_t - \alpha_t| \, \|v_t\|_2 \\
&\le
V\|\hat{\alpha} - \alpha\|_1 \\
&\le
\frac{2VQE}{\sqrt{d_h}}.
\end{aligned}
\]

To connect this to final predictions, suppose the remaining network from the attention output to the final logits is $L$-Lipschitz. Let $\ell$ and $\hat{\ell}$ denote the corresponding final logit vectors. Then
\[
\|\hat{\ell} - \ell\|_\infty
\le
L \|\hat{o} - o\|_2
\le
\frac{2LVQE}{\sqrt{d_h}}.
\]
This immediately bounds the change in output probabilities:
\[
\|\mathrm{softmax}(\hat{\ell}) - \mathrm{softmax}(\ell)\|_1
\le
2\|\hat{\ell} - \ell\|_\infty,
\]
and therefore also controls the per-step increase in negative log-likelihood up to the same order. Over a generation horizon of length $H$, the cumulative degradation scales with $\sum_{i=1}^H E_i$, where $E_i$ is the maximum compressed-token residual at decoding step $i$.

This argument clarifies the role of our two design choices. First, the hybrid storage policy explicitly caps the worst-case residual by retaining high-error tokens in full rank, thereby reducing the effective $E_i$. Second, under standard non-stationary streaming PCA conditions, such as a non-trivial spectral gap and bounded covariance drift, classical analyses of Oja-style subspace tracking imply that the learned subspace continues to track the evolving principal subspace, which helps control residual growth over time. Together, these two mechanisms explain why OjaKV remains more stable than static low-rank compression in long-context generation.

\subsection{Low rank subspace initialization}
\label{sec:low rank init}
We describe here the detailed procedure for constructing the initial projection bases.

For attention head $i$, we gather per-token activations of queries, keys, and values from $n_s$ sampled sequences of length $n$:
\begin{align*}
\mathbf R_i^Q = \bigl[(\mathbf Q_i^1)^\top, \dots, (\mathbf Q_i^{n_s})^\top \bigr] \\
\mathbf R_i^K = \bigl[(\mathbf K_i^1)^\top, \dots, (\mathbf K_i^{n_s})^\top \bigr] \\ 
\mathbf R_i^V = \bigl[(\mathbf V_i^1)^\top, \dots, (\mathbf V_i^{n_s})^\top \bigr]
\end{align*}
where each $\mathbf R_i^{(\cdot)} \in \mathbb R^{(n_s \cdot n)\times d_h}$ and $d_h$ is the head dimension.  

To encourage a shared representation, we concatenate the query and key matrices:
\[
\mathbf R_i^{KQ} = [\,\mathbf R_i^Q, \mathbf R_i^K\,] \in \mathbb R^{(n_s \cdot n)\times 2d_h}.
\]

Applying compact SVD gives
\[
\mathbf R_i^{KQ} = \bm{U} \bm{\Sigma} \bm{V^\top},
\]
with singular values $\sigma_1 \geq \dots \geq \sigma_{d_h}$.  

We select the smallest rank $r$ satisfying the energy criterion
\[
\frac{\|(\mathbf R_i^{KQ})_r\|_F^2}{\|\mathbf R_i^{KQ}\|_F^2} \;\ge\; \epsilon_{\text{th}}.
\]
The top-$r$ columns of $\bm{U}$ define the query–key basis $\bm{U_{k}} \in \mathbb R^{d_h \times r_k}$.  

For the values, we apply the same procedure directly to $\mathbf R_i^V$ to obtain $\bm{U_v} \in \mathbb R^{d_h \times r_v}$.  
Finally, to maintain consistency across attention heads in a layer, we set the effective rank to the maximum $r$ observed in that layer.

\subsection{Equivalence to Full-Rank FlashAttention and Cost Comparison}
\label{sec:eq-cost}
\paragraph{Notation}
Let $\bm{Q}\in\mathbb{R}^{m\times d_h}$ and $\bm{K},\bm{V}\in\mathbb{R}^{n\times d_h}$ denote the per-head query, key, and value blocks, where $m$ is the number of current queries and $n$ is the number of cached keys/values. At the prefilling stage, $m=n$; at the decoding stage, $m=1$.
Let $\bm{U}_{k}\in\mathbb{R}^{d_h\times r_{k}}$ and $\bm{U}_{v}\in\mathbb{R}^{d_h\times r_v}$ be orthonormal bases with $\bm{U}_{k}^{\mathsf{T}}\bm{U}_{k}=\bm{I}_{r_{k}}$ and $\bm{U}_{v}^{\mathsf{T}}\bm{U}_{v}=\bm{I}_{r_v}$.
Define compressed features
\begin{align*}
\bm{\tilde Q} = \bm{Q}\bm{U}_{k}\in\mathbb{R}^{m\times r_{k}}\\
\bm{\tilde K} = \bm{K}\bm{U}_{k}\in\mathbb{R}^{n\times r_{k}}\\
\bm{\tilde V} = \bm{V}\bm{U}_{v}\in\mathbb{R}^{n\times r_v}
\end{align*}
\subsubsection{Equivalence of two computation regimes}
We compare (a) computing attention in the reduced space and expanding the output, versus (b) reconstructing full-rank $\bm{K},\bm{V}$ and calling a standard FlashAttention kernel.
\paragraph{Low-rank kernel (compute-then-expand).}
Form reduced logits and outputs
\begin{align*}
\bm{\tilde{A}} \;=\; \operatorname{softmax}\!\Bigl(\tfrac{\bm{\tilde Q}\,\bm{\tilde K}^{\mathsf{T}}}{\sqrt{d_h}}\Bigr)\in\mathbb{R}^{m\times n} \\
\bm{\tilde O} \;=\; \bm{\tilde{A}}\,\bm{\tilde V} \in\mathbb{R}^{m\times r_v},
\end{align*}
then expand $\bm{\hat O} = \bm{\tilde O}\,\bm{U}_{v}^{\mathsf{T}}\in\mathbb{R}^{m\times d_h}$.
\paragraph{FlashAttention-compatible (reconstruct-then-compute).}
Reconstruct full-rank tensors
\[
\bm{\hat K} = \bm{\tilde K}\,\bm{U}_{k}^{\mathsf{T}}\in\mathbb{R}^{n\times d_h},\qquad
\bm{\hat V} = \bm{\tilde V}\,\bm{U}_{v}^{\mathsf{T}}\in\mathbb{R}^{n\times d_h},
\]
and call FlashAttention with the original queries $\bm{Q}$:
\[
\bm{\hat O} \;=\;
\operatorname{softmax}\!\Bigl(\tfrac{\bm{Q}\,\bm{\hat K}^{\mathsf{T}}}{\sqrt{d_h}}\Bigr)\,\bm{\hat V}
\;\in\;\mathbb{R}^{m\times d_h}.
\]
\paragraph{Lemma (logit equivalence).}
With the above definitions,
\[
\bm{Q}\,\bm{\hat K}^{\mathsf{T}} \;=\; \bm{\tilde Q}\,\bm{\tilde K}^{\mathsf{T}}.
\]
\emph{Proof.}
Since $\bm{\hat K}^{\mathsf{T}}=(\bm{\tilde K}\,\bm{U}_{k}^{\mathsf{T}})^{\mathsf{T}} = \bm{U}_{k}\,\bm{\tilde K}^{\mathsf{T}}$, we have
$\bm{Q}\,\bm{\hat K}^{\mathsf{T}}=\bm{Q}\,(\bm{U}_{k}\,\bm{\tilde K}^{\mathsf{T}})=(\bm{Q}\,\bm{U}_{k})\,\bm{\tilde K}^{\mathsf{T}}=\bm{\tilde Q}\,\bm{\tilde K}^{\mathsf{T}}$.
\paragraph{Corollary (output equivalence).}
The two computation regimes produce the same output $\bm{\hat O}$.
\emph{Proof.}
Starting from the FlashAttention-compatible definition of $\bm{\hat O}$:
\begin{align}
\bm{\hat O} &= \operatorname{softmax}\!\Bigl(\tfrac{\bm{Q}\,\bm{\hat K}^{\mathsf{T}}}{\sqrt{d_h}}\Bigr)\,\bm{\hat V} \tag{FA-compatible} \\
&= \operatorname{softmax}\!\Bigl(\tfrac{\bm{\tilde Q}\,\bm{\tilde K}^{\mathsf{T}}}{\sqrt{d_h}}\Bigr)\,\bm{\hat V} \tag{logit equivalence} \\
&= \operatorname{softmax}\!\Bigl(\tfrac{\bm{\tilde Q}\,\bm{\tilde K}^{\mathsf{T}}}{\sqrt{d_h}}\Bigr)(\bm{\tilde V}\bm{U}_{v}^{\mathsf{T}}) \tag{substitute $\bm{\hat V}$} \\
&= \Bigl(\operatorname{softmax}\!\Bigl(\tfrac{\bm{\tilde Q}\,\bm{\tilde K}^{\mathsf{T}}}{\sqrt{d_h}}\Bigr)\bm{\tilde V}\Bigr)\bm{U}_{v}^{\mathsf{T}} \tag{associativity} \\
&= \bm{\tilde O}\,\bm{U}_{v}^{\mathsf{T}} \tag{low-rank kernel}
\end{align}
Hence the FlashAttention-compatible path is numerically equivalent to computing in the reduced space and then expanding, provided the same scaling $1/\sqrt{d_h}$ is used. Using $1/\sqrt{r_{k}}$ changes the effective temperature and usually needs calibration.
\subsubsection{Complexity and memory comparison}
We summarize per-head costs for a single block with $m$ queries against $n$ cached keys/values. Big-O ignores softmax and masking; General matrix multiply (GEMM) shapes are shown for clarity.
\begin{table*}[t]
\centering
\caption{Computational complexity comparison.}
\renewcommand{\arraystretch}{1.2}
\begin{tabular}{l l l}
\toprule
Regime & Main computations & KV memory \\
\midrule
Full-rank baseline &
\begin{minipage}[t]{0.45\linewidth}
\small
$\bm{Q}\bm{K}^{\mathsf{T}}$: $O(m n d_h)$; \quad $\bm{A}\bm{V}$: $O(m n d_h)$
\end{minipage}
& $2d_h$ \\[4pt]
Low-rank kernel &
\begin{minipage}[t]{0.45\linewidth}
\small
$\bm{\tilde Q}\bm{\tilde K}^{\mathsf{T}}$: $O(m n r_{k})$; \quad $\bm{\tilde{A}}\bm{\tilde V}$: $O(m n r_v)$\\
Expand: $O(m r_v d_h)$
\end{minipage}
& $r_{k}+r_v$ \\[6pt]
FA-compatible &
\begin{minipage}[t]{0.45\linewidth}
\small
Reconstruct $\bm{K},\bm{V}$: $O(n(r_{k}+r_v)d_h)$; \quad FA: $O(m n d_h)$
\end{minipage}
& $r_{k}+r_v$ \\
\bottomrule
\end{tabular}
\end{table*}
\paragraph{Discussion.}
The low-rank kernel reduces the quadratic dot-product costs from $O(m n d_h)$ to $O(m n r_{k})$ and $O(m n r_v)$, plus a linear expansion cost of $O(m r_v d_h)$. The FA-compatible path keeps the full-rank kernel complexity $O(m n d_h)$ but preserves memory savings by storing only $\bm{\tilde K},\bm{\tilde V}$; the reconstruction GEMMs are linear in $n$.
\paragraph{KV-cache memory in bytes.}
Let $b$ be bytes per scalar (e.g., $b{=}2$ for float16). For $L$ layers and $H_{\mathrm{kv}}$ KV heads, the total KV memory for a sequence of length $T$ and batch size $B$ is
\begin{align*}
\text{Full rank:}\quad M_{\text{full}} \;=\; B\,T\,L\,H_{\mathrm{kv}}\,(2d_h)\,b,
\\
\text{Low rank:}\quad M_{\text{low}} \;=\; B\,T\,L\,H_{\mathrm{kv}}\,(r_{k}+r_v)\,b,
\end{align*}
with fractional saving
\[
\text{Saving} \;=\; 1 - \frac{r_{k}+r_v}{2d_h}.
\]
When $r_{k}{=}r_v{=}r$, this simplifies to $\text{Saving}= 1 - \tfrac{r}{d_h}$.

\subsection{Detailed Experimental Setup}
\label{sec:experiment setup}
This section provides a detailed overview of the experimental environment, models, datasets, and evaluation protocols used in this study to ensure full reproducibility of our results.

\textbf{Hardware and Software Environment.}
All experiments were conducted on a single \textbf{NVIDIA H100 NVL} GPU. The software stack was built upon PyTorch and Hugging Face Transformers. The specific versions of the core libraries were as follows: \textbf{PyTorch} \texttt{torch==2.6.0}, \textbf{Transformers} \texttt{transformers==4.44.0}, and \textbf{FlashAttention} \texttt{flash\_attn==2.7.4.post1}. All models were run using their standard float16 precision implementation.

\textbf{Models.}
We evaluated our method on several prominent open-source Large Language Models. For clarity and reproducibility, the specific Hugging Face repository identifiers for each model were: \textbf{Llama-2-7B} (\texttt{meta-llama/Llama-2-7b-chat-hf}), \textbf{Llama-3.1-8B} (\texttt{meta-llama/Llama-3.1-8B-Instruct}), and \textbf{LongChat-7B for RULER} (\texttt{lmsys/longchat-7b-v1.5-32k}).

\textbf{Calibration Dataset.}
The initial low-rank projection bases, $U_{k}$ and $U_{v}$, were derived from a small, general-domain calibration dataset. For this purpose, we used the \textbf{WikiText-2} dataset. The initialization process followed the procedure outlined in Appendix~\ref{sec:low rank init}, where key and value activations were collected from a number of sampled sequences and then decomposed via SVD to form the initial subspaces.

\textbf{Evaluation Benchmarks and Metrics.}
Our comprehensive evaluation was performed across three diverse benchmarks: \textbf{lm-eval-harness}, \textbf{LongBench}, and \textbf{RULER}. Performance was assessed based on the following metrics. \textbf{Accuracy}: We report the specific accuracy metrics as defined by each benchmark's protocol. For lm-eval-harness, this includes the zero-shot accuracy on tasks like PiQA, WinoGrande, and HellaSwag. For LongBench and RULER, this corresponds to their respective scoring mechanisms for long-context reasoning and retrieval tasks. \textbf{GPU Memory}: Memory consumption is reported in Gigabytes (GB) and reflects the specific GPU memory allocated to KV cache during the inference process for a given context length. This provides a practical measure of the hardware requirements. \textbf{Latency}: Latency is reported as Time To First Token (TTFT) in milliseconds (ms), which primarily measures the overhead during the prompt processing (prefill) stage. This is a critical metric for user-facing applications where initial response time is important.

\subsection{Lm-eval-harness}
\label{subsec:lmeval}

To gauge the impact of KV cache compression on downstream utility, we follow the lm-eval-harness protocol on five diverse zero-shot benchmarks: \emph{PiQA}, \emph{WinoGrande}, \emph{ARC-Easy}, \emph{ARC-Challenge}, and \emph{HellaSwag}. We evaluate all methods on the relatively short-context tasks within the lm-eval-harness, where sequence lengths are typically limited to a few hundred tokens.\ As shown in Table~\ref{tab:main_results}, we make two key observations in this setting.\ First, the Eigen-N and StaticPCA baselines yield identical results.\ This finding empirically validates our analysis in Appendix~\ref{sec:eq-cost}, confirming the numerical equivalence between the native low-rank kernel and our FlashAttention-compatible implementation.\ Second, while \kv{} achieves accuracy very close to the full-rank baseline, StaticPCA-H also performs very well, suggesting that the impact of keeping a full rank attention sink and recent window, has a significant impact for these short context-tasks, as these key values contribute to a larger percentage of the total KV cache. Shorter contexts are also more robust to compression, as there is minimal accuracy drop at $0.6\times$ compression for both Llama-2-7B and Llama-3.1-8B. 

Overall, our experiments on these three different benchmarks demonstrate the versatility of \kv{}, and shows that it can perform best in scenarios with long, dynamic context like in RULER.

\begin{table*}[t!]
\small 
\caption{Accuracy (\%) on lm-eval-harness tasks for Llama-2-7B and Llama-3.1-8B.} \label{tab:main_results}
\vspace{-10pt}
\label{tab:main_results}
\centering
\begin{adjustbox}{width=1\textwidth}
\begin{tabular}{@{\extracolsep{6pt}}ccccccccc}
\hline
\multirow{2}{*}{Model}    & \multirow{2}{*}{Compression Ratio} & \multirow{2}{*}{Method}      & \multicolumn{6}{c}{Acc $\uparrow$}                                      \\  \cline{4-9}
                          &                             &                                     & WinoG  & PiQA & HellaS & Arc-e & Arc-c &   Avg-Acc          \\ \hline
\multirow{9}{*}{Llama-2-7B} & \multirow{1}{*}{Full}      & Baseline       & 66.38 & 76.39 & 57.80  & 73.86 & 44.20 & 63.73 \\ \cline{2-9} 
                         & \multirow{3}{*}{0.8x}      & Eigen-N   & 65.90 & 75.03 & 56.29  & 71.30 & 41.38 & 61.98 \\
                         &                            & StaticPCA      & 65.90 & 75.03 & 56.29  & 71.30 & 41.38 & 61.98 \\  
                
                         &                            & \textbf{\kv{}}          & \textbf{66.30} & 75.79 & 57.51  & \textbf{73.86} & \textbf{44.37} & 63.57 \\ \cline{2-9} 
                         & \multirow{3}{*}{0.6x}      & Eigen-N   & 62.98 & 74.54 & 54.99  & 70.03 & 39.93 & 60.49 \\
                         &                            & StaticPCA      & 62.98 & 74.54 & 54.99  & 70.03 & 39.93 & 60.49 \\
                          
                         &                            & \textbf{\kv{}}         & \textbf{66.30} & \textbf{76.39} & \textbf{57.06}  & \textbf{73.74} & \textbf{44.37} & \textbf{63.57} \\ \hline

\multirow{9}{*}{Llama-3.1-8B} & \multirow{1}{*}{Full}      & Baseline       & 73.88 & 80.03 & 59.04  & 81.86 & 51.88 & 69.34 \\ \cline{2-9} 
                           & \multirow{3}{*}{0.8x}      & Eigen-N   & 73.56 & 79.65 & 57.74  & 81.65 & 51.54 & 68.83 \\
                           &                            & StaticPCA      & 73.56 & 79.65 & 57.74  & 81.65 & 51.54 & 68.83 \\
                          
                           &                            & \textbf{\kv{}}          & \textbf{73.95} & \textbf{79.92} & \textbf{59.14}  & \textbf{81.90}  & \textbf{51.79} & \textbf{69.34} \\ \cline{2-9} 
                           & \multirow{3}{*}{0.6x}      & Eigen-N   & 69.85 & 78.67 & 54.87  & 79.38 & 48.12 & 66.18 \\
                           &                            & StaticPCA      & 69.85 & 78.67 & 54.87  & 79.38 & 48.12 & 66.18 \\
                          
                           &                            & \textbf{\kv{}}          & \textbf{73.95} & \textbf{79.82} & \textbf{58.35}  & \textbf{81.90}  & \textbf{51.88} & \textbf{69.18} \\ \hline
\end{tabular}
\end{adjustbox}
\end{table*}

\subsection{Qualitative Analysis and Case Studies}
\label{sec:case study}
To complement the quantitative results presented in the main paper, this section provides a qualitative case study. Our goal is to illustrate the practical impact of OjaKV's online subspace adaptation on generation quality, particularly in long-context scenarios where the distribution of activations can shift significantly. We use a long-document summarization task from the MultiNews subset of the LongBench benchmark, comparing summaries generated by our full OjaKV framework against the StaticPCA baseline at an aggressive 0.6x compression ratio.

\begin{casestudybox}[frametitle={Input Document Snippet (Conceptual Summary)}]
The input is a lengthy (over. 8k tokens) article. It is a multi-doc summarization that requires summarizing over multiple news.
\end{casestudybox}

\begin{casestudybox}[frametitle={Reference Summary}]
The Occupy LA encampment got a stay of execution this morning—though a few arrests have been reported beyond the boundaries of City Hall Park. Two hours after protesters defied city hall's order to vacate the park by 12:01am, lawyer and Occupy supporter Jim Lafferty told Reuters he expected police to give the camp a two-day reprieve. At around 5am, police told protesters over a loudspeaker that they didn't intend to clear the park, the LA Times reports. Police do, however, intend to clear protesters out of the street, and several people who refused to move out of an intersection have already been arrested. \"We definitely need to get morning traffic in,\" a police commander explained. Everything also seems calm at Occupy Philadelphia, where protesters defied a 5pm Sunday deadline to disperse. But as of last night, a police inspector said they didn't intend to forcibly evict anyone. \"We look forward to working with Occupy Philadelphia and a resolution of the problem,\" he told the Philadelphia Inquirer. Confrontation is never good.
\end{casestudybox}

\begin{casestudybox}[frametitle={Baseline: StaticPCA (0.6x Compression)}]
The Occupy Wall Street movement has been making headlines in various cities across the United States. In Philadelphia, Occupy Philly protesters remained on site at City Hall into the evening of November 28, despite a deadline set by Mayor Nutter for them to leave. The protesters, who had set up camp on the City Hall lawn, were given a deadline to vacate the area, but they refused to leave. The police had no plans to evict anyone, and the protesters were allowed to stay on the lawn.
\end{casestudybox}

\textbf{Analysis of Baseline:} The summary generated by StaticPCA exhibits a classic failure mode in long-context summarization: \textbf{topical fixation and critical information loss.} The source document clearly discusses events in two distinct locations, with a primary focus on the tense standoff in Los Angeles and a secondary mention of the calm situation in Philadelphia. However, the StaticPCA summary focuses \textit{exclusively} on Philadelphia, completely omitting any mention of the events in Los Angeles, which constituted the main subject of the report. This suggests that the static principal components, established early, failed to represent the new, dominant information introduced later. The model's inability to adapt to this semantic shift resulted in a factually incomplete and highly misleading summary.

\vspace{1em} 

\begin{casestudybox}[frametitle={OjaKV (0.6x Compression)}]
The Occupy Wall Street movement continues to gain momentum across the United States, with protests taking place in various cities. In Los Angeles, the Occupy LA encampment at City Hall Park was the site of a tense standoff between protesters and police, with commanders later announcing they would not clear the encampment. In Philadelphia, the Occupy Philadelphia campers also remained in place despite a 5 p.m. deadline to vacate the area. The movement is a national one protesting economic inequality, and the response from law enforcement has varied by city.
\end{casestudybox}

\textbf{Analysis of OjaKV:} In stark contrast, the summary from OjaKV successfully captures the multi-faceted nature of the source document. It correctly identifies and synthesizes the key events from \textbf{both Los Angeles and Philadelphia}, presenting a coherent and comprehensive overview. This demonstrates the effectiveness of OjaKV's online subspace adaptation. As the model processed the document and encountered new information related to the LA protest, it dynamically updated its KV cache's principal components. This adaptation allowed it to preserve the crucial details from different sections of the long-context input, avoiding the catastrophic information loss seen in the StaticPCA baseline. The resulting summary is significantly more accurate and useful.

\subsection{Default hyperparameters}
\label{sec:hyper parameters}
\begin{table}[!h]
\centering
\small
\caption{Default hyperparameters unless stated otherwise.}
\begin{tabular}{lcl}
\toprule
Symbol & Default & Description \\
\midrule
$\eta$ & 0.10 & Oja update lr \\
$T$ & 32 & decode update period (steps) \\
$n$ & -- & prompt length \\
$w$ & 32 & importance window size (queries) \\
\bottomrule
\end{tabular}
\end{table}

\subsection{Compatibility with Sequence Length Compression}
\label{sec:token selection}

Our \kv{} method compresses the \emph{feature dimension} ($d \to r$) of the key and value vectors. As a result, it is orthogonal and compatible with sequence length compression techniques the \emph{sequence length} ($n \to m$) such as token eviction or selection.\ This compatibility allows their benefits to be compounded for multiplicative savings.\ We briefly analyze this property theoretically here and validate it experimentally in Appendix~\ref{sec:token selection}.

A token eviction policy can be represented by a selector matrix $\mathbf{S} \in \mathbb{R}^{n \times m}$.\ For a fixed, linear selector, our low-rank projection $\mathbf{U}^\top$ associates perfectly with the selection operation:
\[
  \mathbf U_k^{\top}(\mathbf K\mathbf S)=(\mathbf U_k^{\top}\mathbf K)\mathbf S,
  \quad
  \mathbf U_v^{\top}(\mathbf V\mathbf S)=(\mathbf U_v^{\top}\mathbf V)\mathbf S.
\]
For advanced, context-dependent selectors where $\mathbf{S}_t = \operatorname{Sel}(\mathbf K,\mathbf Q)$ is a function that selects the most relevant tokens based on the current query $\mathbf Q$, the commutation is not exact, but the additional projection error is bounded by the error of the selection policy itself:
\begin{align}
\bigl\| \bm{U}^{\mathsf{T}}\bm{K} - \bm{U}^{\mathsf{T}}\bm{K}\mathbf{S}_t \bigr\|_F
&\le \bigl\| \bm{K} - \bm{K}\mathbf{S}_t \bigr\|_F \nonumber\\
&= \bigl\|\mathbf K-\operatorname{Sel}(\mathbf K,\mathbf Q)\bigr\|_F .
\end{align}
because left-multiplication by $\bm{U}^{\mathsf{T}}$ is a contraction with respect to the Frobenius norm. This operational orthogonality means that combining a rank-$r$ \kv{} with a policy retaining $m$ of $n$ tokens yields a total compression ratio of $\text{CR} = (d/r) \times (n/m)$.


We also provides empirical validation for this claim by combining OjaKV with SnapKV~\citep{li2024snapkv}, a representative token selection method.

\begin{table*}[h!]
\centering
\small
\caption{Compounded KV cache compression by combining OjaKV with SnapKV. The total compression ratio demonstrates multiplicative savings, offering a compelling trade-off between performance and memory efficiency.}
\label{tab:compounded_compression}
\begin{tabular}{@{}lcccc@{}}
\toprule
\textbf{Method} & \textbf{Rank Comp.} & \textbf{Token Keep Rate} & \textbf{Memory Usage (\%)} & \textbf{Accuracy} \\
\midrule
Full KV Cache (Baseline) & 1.0x & 100\% & 100\% &  53.0\\
\midrule
SnapKV (Token Sel. only) & 1.0x & 50\% & 50\% & 52.66\\
OjaKV (Rank Comp. only) & 1.67x (0.6x) & 100\% & 60\% & 43.13\\
\textbf{OjaKV + SnapKV} & \textbf{1.67x (0.6x)} & \textbf{50\%} & \textbf{30\%} & \textbf{43.33} \\
\bottomrule
\end{tabular}
\end{table*}

\textbf{Experimental Setup.} We chose SnapKV as it is a strong baseline that uses importance scores to identify and retain salient tokens. We evaluated four configurations on the LongBench benchmark suite using the Llama-3.1-8B model. The configurations are: (1) the uncompressed baseline, (2) SnapKV alone with a 50\% token keep rate, (3) OjaKV alone with a 0.6x rank compression, and (4) a combined approach applying both OjaKV's rank compression and SnapKV's token eviction. Performance is measured by the average accuracy across LongBench tasks, and efficiency is measured by the total KV cache compression ratio.

\textbf{Results and Analysis.} The results, presented in Table~\ref{tab:compounded_compression}, confirm our hypothesis. Our analysis shows that \textbf{\kv{}} can be effectively combined with token eviction methods like SnapKV. This compounded approach further reduces KV cache memory usage with only a minor, graceful degradation in model accuracy. This result validates that our feature-dimension compression is complementary to sequence-length compression, offering a practical path to even greater memory efficiency.

\end{document}

%% file: Sections/Chapter_1.tex
\vspace{-15pt}
\section{Introduction}
\vspace{-5pt}
Large language models (LLMs) such as GPT-4o~\citep{openai2024openaio1card} and Deepseek-R1~\citep{deepseekai2025deepseekr1incentivizingreasoningcapability} have demonstrated remarkable performance across diverse domains, including coding~\citep{nam2024using}, mathematics~\citep{setlur2024rl}, and open-ended text generation~\citep{kumichev2024medsyn}.\ However, as model capabilities and context length expand, GPU memory emerges as a critical bottleneck for inference.\ The memory footprint arises from two primary sources: (i) model weights, with a model like Llama-3.1-8B requiring 16\,GB alone; and (ii) the Key-Value (KV) cache used during prompt prefilling and autoregressive decoding.\ For instance, processing a 32K-token prompt with Llama-3.1-8B in float16 precision at a batch size of 4 consumes an additional 16\,GB for the KV cache, rivalling the size of the model weights themselves.\ This substantial memory consumption makes long-context inference prohibitive on all but high-end hardware.

To mitigate this challenge, a variety of methods have been proposed to optimize KV-cache memory usage~\citep{shi2024keep}.\ These approaches can be grouped into four categories: (1) \emph{Quantization}, which stores keys and values at a lower precision (e.g., 8-bit)~\citep{hooper2024kvquant, liu2024kivi}; (2) \emph{Token Selection}, which prunes or merges tokens deemed unimportant based on attention scores or heuristic saliency measures~\citep{xiao2023efficient, li2024snapkv, zhang2023h2o}; (3) \emph{Offloading}, which transfers the KV cache to CPU memory and selectively streams it back during decoding~\citep{tang2024quest, sun2024shadowkv, zhu2025sentencekv}; and (4) \emph{Low-rank Approximation}, which projects keys and values into a lower-dimensional subspace~\citep{saxena2024eigen, lin2024matryoshkakv}.

Our work focuses on this fourth direction.\ By compressing each key and value vector from dimension $d$ (e.g., $d=128$) to $r$ (e.g., $r=96$), low-rank methods can reduce cache memory by $((1-r/d) \times 100)\%$ while preserving model accuracy, yielding substantial savings for long-context inference.\ While token selection has become a widely adopted strategy, we provide theoretical support showing that low-rank projection and token eviction are compatible, offering multiplicative benefits that enable even greater memory reductions when combined.

Existing low-rank methods fall into two main categories.\ (1) \emph{Weight-decomposition} techniques directly factorize the linear projection weights for query, key, and value ($\bm{W}_q$, $\bm{W}_k$, $\bm{W}_v$) into low-rank matrices, thereby caching already-compressed intermediate states~\citep{chang2024palucompressingkvcachelowrank}.\ However, this approach often incurs a noticeable degradation in accuracy.\ (2) \emph{Projection-based} techniques learn fixed orthonormal projeciton bases ($\bm{U}_q$, $\bm{U}_k$, $\bm{U}_v$) from a calibration dataset.\ These bases are then used to compress the KV cache, which is reconstructed during attention computation~\citep{saxena2024eigen, lin2024matryoshkakv}.\ While effective, these static bases implicitly assume that inference prompts will follow the same distribution as the calibration data.\ In practice, distribution shifts (e.g., from dialogue to code generation) cause the approximation to deteriorate, harming generation quality.


To address these limitations, we propose \textbf{OjaKV}, a novel framework for KV-cache compression that operates on two core principles.\ Our first key insight is that uniform compression across all tokens is suboptimal.\ Motivated by the findings of attention sinks~\citep{xiao2023efficient} and SnapKV~\cite{li2024snapkv}, OjaKV employs a hybrid storage policy that strategically excludes the crucial tokens with high reconstruction error from low-rank projection.\ This preserves their full-rank fidelity, creating stable anchors for the attention mechanism and forming a significantly stronger performance baseline.\ Second, for the remaining intermediate tokens, we incorporate online subspace adaptation using Oja's incremental principle component analysis (PCA)~\citep{oja1997nonlinear}.\ This mechanism performs a comprehensive update during the prefill stage on a selection of salient tokens, and subsequently executes periodic lightweight updates during decoding.\ This ensures the low-rank basis continuously adapts to the evolving context with negligible overhead.\ Our framework is fully compatible with modern attention modules such as FlashAttention~\citep{dao2022flashattentionfastmemoryefficientexact}, ensuring practicality in real-world long-context inference.

We evaluate \kv~on multiple-choice benchmarks from the lm-eval-harness~\citep{biderman2024lessons}, aligning with prior studies~\citep{saxena2024eigen, lin2024matryoshkakv}.\ Additionally, for the first time to the best of our knowledge, we evaluate projection-based low-rank KV cache compression methods' performance on generation-centric long-context tasks using LongBench~\citep{bai2023longbench} and RULER~\citep{hsieh2024ruler}.\ We also evaluate our method on long reasoning and generation task, AIME~\cite{balunovic_srimatharena_2025}. Across all settings, \kv~demonstrates superior performance over static low-rank baselines at the equivalent compression ratios.

In summary, our contributions are threefold.\ First, we introduce a hybrid low-rank KV-cache compression framework, OjaKV, which combines a selective full-rank storage policy with a context-aware online subspace adaptation.\ Second, our design is compatible with FlashAttention, ensuring practicality for modern long-context inference pipelines.\ Third, we conduct the first comprehensive evaluation of low-rank KV compression on challenging generation-centric benchmarks, moving beyond the simpler multiple-choice tasks.

%% file: Sections/Chapter_2.tex
\section{Related Work}
Recent work on reducing the memory footprint of LLM inference has led to various KV cache compression strategies, including quantization, token pruning, offloading, and subspace compression.\ Our method builds on low-rank approximation, extending it with an online, context-adaptive formulation.\ We review relevant methods below.\

\subsection{KV-Cache Compression}

The memory overhead of the KV cache motivates four main compression strategies~\citep{shi2024keep}.\
(1) \emph{Quantization}: These methods reduce precision (e.g., to 4-bit or 2-bit) for storage.\ Approaches like KVQuant~\citep{hooper2024kvquant} and KIVI~\citep{liu2024kivi} achieve significant compression with minimal quality loss.\
(2) \emph{Token selection}: This category retains only essential tokens.\ StreamingLLM~\citep{xiao2023efficient} keeps ``attention sinks,'' while SnapKV~\citep{li2024snapkv} selects tokens based on importance scores.\
(3) \emph{Offloading}: Methods such as Quest~\citep{tang2024quest} and ShadowKV~\citep{sun2024shadowkv} store the cache in CPU memory, selectively reloading relevant parts during computation.\
(4) \emph{Low-rank approximation}: These approaches project keys and values into a lower-dimensional subspace~\citep{saxena2024eigen, lin2024matryoshkakv}.\
Our method falls into this last category and is \emph{orthogonal} to the others, enabling additive memory savings.\ Appendix~\ref{sec:token selection} provides analysis and experiments showing that our method can be combined with token-eviction-based methods.

\vspace{-5pt}
\subsection{Low-Rank Approximation for Attention}

Low-rank structures are widely utilized for compression.\ Palu~\citep{chang2024palucompressingkvcachelowrank} and ReCalKV~\citep{yan2025recalkv} factorize weights to cache compressed states, though often with accuracy degradation.\
EigenAttention~\citep{saxena2024eigen} and MatryoshkaKV~\citep{lin2024matryoshkakv} project activations using bases derived from calibration data ($\bm{U_k}, \bm{U_v}$).\ However, these methods rely on a \emph{static} basis, which degrades performance under distribution shifts.\
While architectures like Multi-Head Latent Attention~\citep{liu2024deepseek, liu2024deepseek1} integrate low-rank structures natively, they require training new models.\
Our method addresses these limitations via \emph{online subspace adaptation}, continuously updating projection matrices during inference for a plug-in, context-aware approximation.\

\vspace{-5pt}
\subsection{Online Principal Component Analysis}

In autoregressive decoding, rerunning SVD on the entire history is computationally infeasible.\ Online PCA algorithms address this by incrementally updating the subspace.\
Perturbation techniques~\citep{gu1994stable} provide accuracy but are too resource-intensive for high-dimensional streams.\
Incremental SVD~\citep{brand2002incremental} offers a middle ground but remains costly for real-time LLM inference.\
In contrast, stochastic methods like Oja's rule~\citep{oja1997nonlinear} efficiently optimize variance via gradient updates.\
Our proposed method, OjaKV, builds on this foundation.\ By applying Oja's rule to update key and value subspaces on-the-fly, it enables fast, adaptive, and calibration-free approximation.\

%% file: Sections/Chapter_3.tex
\section{Preliminaries: Low-Rank Attention}

We begin by describing the standard attention mechanism and how it can be adapted for low-rank KV cache compression. For simplicity, we focus on a single attention head.\ Let the head dimension be $d_h$ and the input sequence be $\bm{X} \in \mathbb{R}^{n \times d}$.\ Standard attention first projects the input into query, key, and value representations using projection matrices $\bm{W}_q$, $\bm{W}_k$, $\bm{W}_v$ $\in \mathbb{R}^{d \times d_h}$:
\[
\bm{Q} = \bm{X}\bm{W}_q, \qquad \bm{K} = \bm{X}\bm{W}_k, \qquad \bm{V} = \bm{X}\bm{W}_v,
\]
where $\bm{Q}, \bm{K}, \bm{V} \in \mathbb{R}^{n \times d_h}$.\ The KV cache stores $\bm{K}$ and $\bm{V}$ for future use during decoding. The attention scores $\bm{A}$ are then computed as the scaled dot product between queries and keys, followed by a softmax operation. The output of the attention head, $\bm{O}$, is computed by applying these attention scores to value matrix $\bm{V}$:
$
\bm{A} = \text{softmax}\left(\bm{Q}\bm{K}^T/\sqrt{d_h}\right) \in \mathbb{R}^{n \times n}, \bm{O} = \bm{A}\bm{V} \in \mathbb{R}^{n \times d_h}
$.
\subsection{Low-Rank KV-Cache Approximation}
The core idea of low-rank approximation is to project $\bm{K}$ and $\bm{V}$ onto lower-dimensional subspaces.\ We define two orthonormal bases: $\bm{U}_{k} \in \mathbb{R}^{d_h \times r_{k}}$ for keys and queries, and $\bm{U}_{v} \in \mathbb{R}^{d_h \times r_{v}}$ for values, where $r_{k}, r_{v} \ll d_h$ are the desired ranks for compression.\ The bases satisfy $\bm{U}_{k}^{\mathsf{T}}\bm{U}_{k} = \bm{I}_{r_{k}}$ and $\bm{U}_{v}^{\mathsf{T}}\bm{U}_{v} = \bm{I}_{r_{v}}$.\ The low-rank bases $\bm{U}_{k}$ and $\bm{U}_{v}$ are initialized from a small calibration dataset (see Appendix~\ref{sec:low rank init} for details).\ 

Instead of caching the full-rank $\bm{K}$ and $\bm{V}$, we store their \emph{compressed} representations:
\[
\bm{\tilde{K}} = \bm{K}\bm{U}_{k} \in \mathbb{R}^{n \times r_{k}}, \qquad
\bm{\tilde{V}} = \bm{V}\bm{U}_{v} \in \mathbb{R}^{n \times r_{v}}.
\]
This reduces per-token storage requirement for the KV cache from $2d_h$ to $r_{k} + r_{v}$.\ If needed, the full-rank matrices can be approximately reconstructed via $\bm{\hat{K}} = \bm{\tilde{K}}\bm{U}_{k}^{\mathsf{T}}$ and $\bm{\hat{V}} = \bm{\tilde{V}}\bm{U}_{v}^{\mathsf{T}}$.

\subsection{Efficient Attention Computation}
This compressed representation allows for a more efficient attention calculation.\ Projecting queries into the shared query-key subspace yields $\bm{\tilde{Q}} = \bm{Q}\bm{U}_{k}$.\ The attention scores can be computed directly in the low-rank space, as the projection is mathematically equivalent to using the reconstructed matrices:
\begin{equation}
\begin{split}
\bm{\tilde{Q}}\bm{\tilde{K}}^{\mathsf{T}} &= (\bm{Q}\bm{U}_{k})(\bm{K}\bm{U}_{k})^{\mathsf{T}} \\
&= \bm{Q}\bm{U}_{k}\bm{U}_{k}^{\mathsf{T}}\bm{K}^{\mathsf{T}} = \bm{Q}\bm{U}_{k}(\bm{K}\bm{U}_{k})^{\mathsf{T}} \\
&= \bm{Q}\bm{U}_{k}\bm{\tilde{K}}^{\mathsf{T}} \\
&= \bm{Q}\bm{\hat{K}}^{\mathsf{T}} \in \mathbb{R}^{n \times n}
\end{split}
\end{equation}
Thus, attention computed in the low-rank space is equivalent to using the reconstructed keys. The full attention operation is then performed in the low-rank space, and the final output is projected back to the original dimension:
$ 
\bm{\tilde{O}} = \mathrm{softmax}\!\left(\bm{\tilde{Q}}\bm{\tilde{K}}^{\mathsf{T}}/{\sqrt{d_h}}\right) \bm{\tilde{V}} \in \mathbb{R}^{n \times r_v},
\bm{\hat{O}} = \bm{\tilde{O}}\bm{U}_{v}^{\mathsf{T}} \in \mathbb{R}^{n \times d_h}.
$

We note that the final output computation in the above formulation is mathematically equivalent to using the reconstructed keys and values, i.e., $\bm{\hat{O}} = \mathrm{softmax}\!\left(\bm{{Q}}\bm{\hat{K}}^{\mathsf{T}}/{\sqrt{d_h}}\right) \bm{\hat{V}}$.

\subsection{Practical Implementation: Compatibility with FlashAttention}

Optimized attention kernels like FlashAttention operate on full-dimensional tensors of shape $(n \times d_h)$ and cannot directly use the compressed features.\ To maintain compatibility, we store the compressed KV cache ($\bm{\tilde{K}}$, $\bm{\tilde{V}}$) and perform on-the-fly reconstruction before using FlashAttention.\ Given queries ${\bm{Q} \in \mathbb{R}^{n \times d_h}}$, we reconstruct the keys and values in the original space from the compressed cache:
\[
\bm{\hat{K}} = \bm{\tilde{K}}\bm{U}_{k}^{\mathsf{T}} \in \mathbb{R}^{n \times d_h}, \qquad
\bm{\hat{V}} = \bm{\tilde{V}}\bm{U}_{v}^{\mathsf{T}} \in \mathbb{R}^{n \times d_h}.
\]
These reconstructed tensors are then passed to FlashAttention:
\[
\bm{O}_{\text{out}} = \bm{\hat{O}} = \mathrm{FlashAttention}(\bm{Q}, \bm{\hat{K}}, \bm{\hat{V}}).
\]
This approach maintains the memory savings of a compressed cache while incurring only a modest runtime overhead, as detailed in Appendix~\ref{sec:eq-cost}.

\section{Motivation}
\label{sec:motivation}

Offline low-rank bases are fitted to the calibration distribution and can misalign under domain or task shifts at inference, increasing projection error for keys and values.\ We therefore maintain an adaptive basis that periodically refreshing the basis with key and value features from current prompts and generated content the model can track distributional shifts and maintain alignment between the low-rank subspace and the evolving sequence.\ To validate this hypothesis, we conduct an experiment, summarized in Table~\ref{tab:rer_so}.\ We compute an initial basis, $\bm{U}_{\text{cal}}$, from a general-domain corpus (WikiText-2~\citep{guo2020wiki}) and evaluate it on a long-context news summarization task from a different domain (the MultiNews subset of LongBench~\citep{bai2023longbench}).\ We compare this against an adapted basis, $\bm{U}_{\text{adapt}}$, formed by updating $\bm{U}_{\text{cal}}$ online with Oja's rule after processing a short prefix of the MultiNews data.\ An oracle basis, $\bm{U}_{\text{test}}$, computed via PCA on the full test set, serves as an upper bound on performance.



\begin{table}[h]
    \centering
    \small
    \caption{
        RER and SO on the MultiNews test set. Adapting the basis online reduces the projection error (RER) and improves alignment with the oracle test basis ($\bm{U}_{\text{test}}$).
    }
    \label{tab:rer_so}
    \begin{tabular}{lcc}
        \toprule
        Basis & RER (on Test)\,\(\downarrow\) & SO with $\bm{U}_{\text{test}}$\,\(\uparrow\) \\
        \midrule
        $\bm{U}_{\text{cal}}$ & 0.255 & 0.597 \\
        $\bm{U}_{\text{adapt}}$ & 0.097 & 0.653 \\
        \bottomrule
    \end{tabular}
\end{table}
We use two metrics: \textbf{Residual-Energy Ratio (RER)}~\citep{najafzadeh2024residual} measures the projection error, and \textbf{Subspace Overlap (SO)}~\citep{knyazev2012principal}, which we compute against the oracle basis ($\bm{U}_{\text{test}}$) to quantify alignment, defined as $\mathrm{SO}(\bm{U}_1, \bm{U}_2) = \mathrm{Tr}(\bm{U}_1^{\mathsf{T}}\bm{U}_2\bm{U}_2^{\mathsf{T}}\bm{U}_1)/r$. The in-domain RER of $\bm{U}_{\text{cal}}$ on the calibration set is 0.035. As shown in Table~\ref{tab:rer_so}, the static calibration basis generalizes poorly under distribution shift: its RER increases to 0.255 on the new task. Applying a lightweight Oja update to form $\bm{U}_{\text{adapt}}$ mitigates this, lowering the RER to 0.097. This adaptation also improves alignment with the oracle basis, raising the SO from 0.597 to 0.653. These findings confirm that online updates can effectively counteract distribution shift.


\section{Methodology}

\begin{figure*}[!t]
\centering
\includegraphics[width=0.9\textwidth]{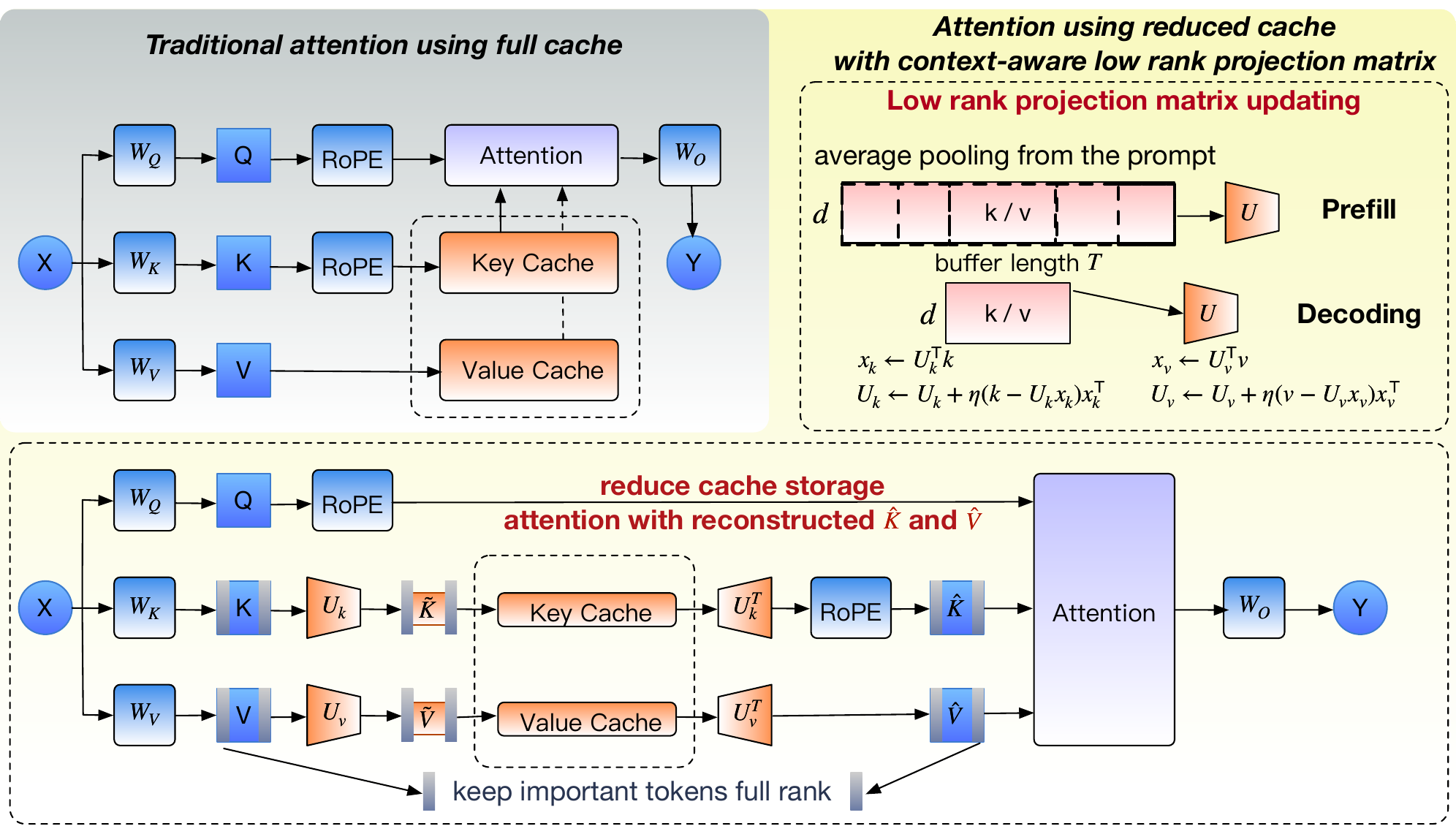}
\caption{Overview of the \kv{} workflow.
    The \textbf{top-left panel} shows standard attention using full-rank KV caching. Our method, shown in the \textbf{bottom panel}, introduces a low-rank path where keys and values are compressed using projection matrices ($\bm{U}_{k}, \bm{U}_v$) before caching.
    The \textbf{top-right inset} illustrates the core mechanism: these projection matrices are dynamically updated during both the prefill and decoding phases to adapt to the context.}
\label{fig2}
\end{figure*}

Our method, OjaKV, introduces a hybrid strategy for memory-efficient inference that combines selective full-rank retention, with the goal of preserving high-fidelity representations for critical tokens, with online-adapted low-rank compression for remaining sequence. As illustrated in Figure \ref{fig2}, most key and value vectors are projected into a compact subspace via learned projection matrices, which are continuously adapted during inference to remain aligned with the evolving context.

Our framework is built around three core components: (i) a hybrid KV cache storage policy that exempts key contextual tokens from compression; (ii) a lightweight initialization procedure that seeds the projection matrices from a small calibration corpus (see Appendix \ref{sec:low rank init} for details), (iii) a two-phase online update scheme using Oja’s algorithm to adapt the low-rank subspace during both the prompt (prefill) and decoding stages (Appendix \ref{subsec: algo}).

\subsection{Hybrid KV Storage with Reconstruction-Error Selection}

In long-context generation, not all tokens contribute equally to downstream predictions, and crucially, not all tokens suffer equally from low-rank approximation error. Our hybrid storage policy dynamically identifies which tokens should retain full-rank representations based on their \emph{reconstruction error} under the current low-rank basis.

\paragraph{Error-Based Token Selection.}
For each key token, we compute the residual between the full-rank key and its low-rank reconstruction:
\[
\bm{r}_t = \bm{k}_t - \bm{U}_k \bm{U}_k^\top \bm{k}_t
\]
This residual directly governs the attention score perturbation: for any query $\bm{q}$, the dot-product error introduced by compression, $\Delta = \bm{q}^\top \bm{r}_t / \sqrt{d}$, is bounded by $|\Delta| \le \|\bm{q}\| \|\bm{r}_t\| / \sqrt{d}$ via Cauchy-Schwarz. Unlike positional heuristics (e.g., retaining initial tokens as ``attention sinks''), which assume reconstruction error is negligible beyond a fixed window, our approach explicitly bounds error by selecting tokens based on their actual residual magnitude.

To assess downstream impact, we weight residuals by attention from recent queries. Let $\bm{Q}_{\text{win}}$ denote queries from a sliding window of the last $w$ positions. We compute query-weighted error scores:
\[
e_t = \frac{1}{|\mathcal{W}_t|} \sum_{h,q \in \mathcal{W}_t} \left| \frac{\bm{q}_{h,q}^\top \bm{r}_t}{\sqrt{d_h}} \right|
\]
where $\mathcal{W}_t$ denotes query positions that can attend to token $t$ under causal masking. After optional pooling, we retain the top-$k$ tokens with highest error scores at full rank while compressing all other tokens. This ensures that tokens poorly represented by the current basis retain their full expressiveness, while the attention perturbation for all compressed tokens remains bounded with theoretical guarantees that static positional policies lack.





\paragraph{From residual error to generation quality.}
Our residual-based selection rule also admits an end-to-end error interpretation. Let $\hat{k}_t = k_t + e_t$ denote the compressed key with residual $e_t$, and assume $\|q\|_2 \le Q$ and $\|v_t\|_2 \le V$. Then each logit perturbation satisfies
\[
|\hat{z}_t - z_t|
=
\frac{|q^\top e_t|}{\sqrt{d_h}}
\le
\frac{Q\|e_t\|_2}{\sqrt{d_h}}.
\]
If $E = \max_t \|e_t\|_2$ over compressed tokens, standard softmax stability gives
\[
\begin{aligned}
\|\mathrm{softmax}(\hat{z}) - \mathrm{softmax}(z)\|_1
&\le
2\|\hat{z} - z\|_\infty \\
&=
O(QE/\sqrt{d_h}).
\end{aligned}
\]
Hence the attention-output error is bounded as
\[
\|\hat{o} - o\|_2
\le
V \|\Delta \alpha\|_1
=
O(VQE/\sqrt{d_h}).
\]
If the remaining network mapping from the attention output to final logits is $L$-Lipschitz, then the final logit perturbation is bounded by
\[
\|\Delta \ell\|_\infty \le L \|\hat{o} - o\|_2,
\]
which in turn controls the output distribution shift and the per-step increase in negative log-likelihood. Therefore, reducing the residual norm of compressed tokens directly improves generation fidelity. Our hybrid policy further limits the worst-case error by retaining high-residual tokens in full rank, while online Oja updates reduce residual growth under distribution shift by continuously adapting the subspace. See Appendix~\ref{sec:theory_guarantee} for the detailed proof.

\subsection{Two-Phase Online Updates with Oja's Algorithm}

The low-rank bases $\bm{U}_k$ and $\bm{U}_v$ are initialized to capture the principal subspaces of key and value representations. Recall that in standard attention, the output for query $\bm{q}_t$ is computed as:
\[
\text{Attn}(\bm{q}_t, \bm{K}, \bm{V}) = \sum_{i \le t} \frac{\exp(\bm{q}_t^\top \bm{k}_i / \sqrt{d})}{\sum_{j \le t} \exp(\bm{q}_t^\top \bm{k}_j / \sqrt{d})} \bm{v}_i
\]
The fidelity of our low-rank approximation depends on how well $\bm{U}_k$ and $\bm{U}_v$ span the subspaces occupied by keys and values, respectively. Formally, the optimal rank-$r$ basis minimizes the expected reconstruction error $\mathbb{E}[\|\bm{x} - \bm{U}\bm{U}^\top \bm{x}\|^2]$, which is achieved by the top-$r$ eigenvectors of the covariance matrix $\bm{C} = \mathbb{E}[\bm{x}\bm{x}^\top]$.

However, during autoregressive generation, newly decoded tokens may introduce key-value patterns that deviate from the initial prompt distribution. A static basis trained only on the prompt will accumulate approximation error as the context grows, degrading attention accuracy. To address this, we update the projection bases online using Oja's rule~\citep{oja1982simplified}, a streaming algorithm that incrementally tracks the principal subspace of non-stationary data.

\paragraph{Oja's Rule for Subspace Tracking.}
Given a new sample $\bm{x}$, Oja's rule updates the basis $\bm{U} \in \mathbb{R}^{d \times r}$ as:
\[
\bm{U} \leftarrow \bm{U} + \eta \left( \bm{x}\bm{x}^\top \bm{U} - \bm{U}\bm{U}^\top \bm{x}\bm{x}^\top \bm{U} \right)
\]
This can be interpreted as stochastic gradient ascent on $\text{tr}(\bm{U}^\top \bm{C} \bm{U})$ with an implicit orthonormality constraint. The first term $\bm{x}\bm{x}^\top \bm{U}$ pulls the basis toward directions of high variance, while the second term $\bm{U}\bm{U}^\top \bm{x}\bm{x}^\top \bm{U}$ enforces approximate orthogonality. For batch updates with samples $\bm{X} \in \mathbb{R}^{n \times d}$, we use the empirical covariance $\bm{C} = \bm{X}^\top \bm{X} / n$.

\paragraph{Prefill Stage.}
During prefill, we initialize the bases using the full set of key and value vectors from the prompt. To reduce redundancy from positionally adjacent tokens and lower computational cost, we apply local average pooling before the update:
\begin{align*}
\bm{X}_{\text{pooled}} &= \text{AvgPool}(\bm{X}, b), \\
\bm{U} &\leftarrow \bm{U} + \eta \left( \bm{C}_{\text{pooled}} \bm{U} - \bm{U} \bm{U}^\top \bm{C}_{\text{pooled}} \bm{U} \right)
\end{align*}
where $\bm{C}_{\text{pooled}} = \bm{X}_{\text{pooled}}^\top \bm{X}_{\text{pooled}} / N$, and $b$ is the pooling window size. After the update, we re-orthonormalize via QR decomposition to ensure numerical stability.

\paragraph{Decoding Stage.}
During decoding, each generated token produces new key-value pairs that may lie outside the current subspace. We accumulate these vectors in a buffer $\mathcal{B}$. Every $T$ steps, we apply Oja's rule to the buffered features:
\[
\bm{U} \leftarrow \bm{U} + \eta \left( \bm{C}_{\mathcal{B}} \bm{U} - \bm{U} \bm{U}^\top \bm{C}_{\mathcal{B}} \bm{U} \right)
\]
with a conservative learning rate to balance adaptation against stability. After the update, we re-orthonormalize the bases and clear the buffer. This periodic refinement ensures that the low-rank subspace continuously tracks the evolving key-value distribution throughout generation.

\paragraph{Practical variant: OjaKV-PF} For fair comparison with prior low-rank baselines such as EigenAttention and StaticPCA, the main experiments in this paper use reconstructed keys and values for attention computation during both prefilling and decoding. In practice, however, the prefilling stage can instead compute attention using the original full-rank keys and values, while still using the prompt features to update the low-rank bases and populate the compressed KV cache. We refer to this practical variant as \textsc{OjaKV-PF}, where ``PF'' denotes full-rank prefilling. After prefilling, decoding continues to use reconstructed keys and values from the compressed cache. This variant preserves the decoding-time memory savings of OjaKV while reducing unnecessary approximation overhead during prefilling.

In the practical variant \textsc{OjaKV-PF}, the prompt key and value features are still used to update the low-rank bases and determine the hybrid storage policy, but the prefilling attention itself is computed using the original full-rank keys and values. Low-rank reconstruction is then used only after prefilling, during autoregressive decoding.

%% file: Sections/Chapter_4.tex
\section{Experiments}

In this section, we present the experimental setup and evaluate \textbf{OjaKV} against relevant baselines in realistic long-context inference scenarios. All experiments are conducted on NVIDIA H100 GPUs using \textbf{Llama-2-7B} and \textbf{Llama-3.1-8B}~\citep{grattafiori2024llama3herdmodels}, evaluated on three diverse benchmarks: \textbf{RULER}~\citep{hsieh2024ruler}, \textbf{LongBench}~\citep{bai2023longbench}, and \textbf{AIME 2025}~\citep{balunovic_srimatharena_2025}. Additional results on \textbf{lm-eval-harness}~\citep{agarwal2024copilotevaluationharnessevaluating} are provided in the Appendix~\ref{subsec:lmeval}.


We compare \textbf{OjaKV} against four baselines: (1)~\textbf{Full KV Cache}, the uncompressed baseline serving as a performance upper bound; (2)~\textbf{Eigen-N}, a direct low-rank implementation of prior work~\citep{saxena2024eigen} that is impractical for long contexts due to its incompatibility with FlashAttention; (3)~\textbf{StaticPCA}, which uses the same fixed, offline-computed SVD basis as Eigen-N but reconstructs full-rank tensors on-the-fly for compatibility with modern inference; and (4)~\textbf{Palu}~\citep{chang2024palucompressingkvcachelowrank}, which decomposes model weights into low-rank matrices. Unless otherwise specified, all projection-based methods, including OjaKV, use reconstructed keys and values during prefilling, which provides a controlled and fair comparison across methods. We also consider a practical variant, \textbf{OjaKV-PF}, which uses full-rank attention during prefilling and reconstructed keys and values only during decoding. We report model accuracy, memory usage (GB), and Time to First Token (TTFT), with detailed configurations in Appendices~\ref{sec:experiment setup} and~\ref{sec:hyper parameters}.


\begin{table*}[t]
\fontsize{18}{24}\selectfont
\setlength{\tabcolsep}{5pt}
\centering
\caption{Accuracy (\%) on LongBench tasks for Llama2-7B and Llama-3.1-8B.}\label{tab:longbench}
\vspace{-10pt}
\begin{threeparttable}
\scalebox{0.38}{
\begin{tabular}{l|lcccccccccccc}
\specialrule{1pt}{0pt}{2pt}
 & \multirow{4}{*}{~~~LLMs {\huge }} & \multicolumn{2}{c}{Single-Document QA} & \multicolumn{3}{c}{Multi-Document QA} & \multicolumn{2}{c}{Few-shot Learning} & \multicolumn{2}{c}{Synthetic} & \multicolumn{2}{c}{Code} & \multicolumn{1}{c}{Avg} \\
\cmidrule(lr){3-4}\cmidrule(lr){5-7}\cmidrule(lr){8-9}\cmidrule(lr){10-11}\cmidrule(lr){12-13}\cmidrule(lr){14-14}
 && \rotatebox[origin=c]{30}{NrtvQA} & \rotatebox[origin=c]{30}{MF-en} & \rotatebox[origin=c]{30}{HotpotQA} & \rotatebox[origin=c]{30}{2WikiMQA} & \rotatebox[origin=c]{30}{Musique} & \rotatebox[origin=c]{30}{TREC} & \rotatebox[origin=c]{30}{SAMSum} & \rotatebox[origin=c]{30}{PCount} & \rotatebox[origin=c]{30}{PRe} & \rotatebox[origin=c]{30}{Lcc} & \rotatebox[origin=c]{30}{RB-P} & \rotatebox[origin=c]{30}{Avg} \\
\specialrule{1pt}{2pt}{2pt}
\multirow{5}{*}{\rotatebox[origin=c]{90}{\fontsize{15}{100}\selectfont Llama-2-7B}}
&~~~Full KV   & 18.79 & 34.41 & 25.3 & 28.33 & 8.52 & 0.0 & 6.22 & 1.55 & 9.0 & 15.08 & 17.35 & \avgXI{18.79+34.41+25.3+28.33+8.52+0.0+6.22+1.55+9.0+15.08+17.35} \\
\cline{2-14}
&~~~Eigen-N 0.8x & OOM & OOM & OOM & OOM & OOM & OOM & OOM & OOM & OOM & OOM & OOM & NA\\
&~~~StaticPCA 0.8x & 16.95       & 32.8             & 21.31    & 24.73    & 6.28    & 0    & 5.34   & 2.25           & 2.61                   & 14.06 & 16.78       & 13.01 \\
&~~~Palu 0.8x  & 15.75       & 21.32            & 8.5      & 10.11    & 4.34    & 0    & 4.14   & 2.35           & 6.58                   & 6.48  & 5.31        & 7.72  \\
&~~~\textbf{\kv{}}~0.8x & 16.28       & 31.19            & 20.4     & 23.7     & 7.44    & 0    & 3.91   & 1.61           & 4                      & 16.16 & 20.85       & 13.23 \\

&~~~\textit{OjaKV-PF}~0.8x   &  16.77       & 33.62            & 23.3    &  26.38   & 7.33   & 0  & 5.28   & 4            & 1.5                    & 16.05 & 17.48       & 13.79 \\

\cline{2-14}
&~~~Eigen-N 0.6x & OOM & OOM & OOM & OOM & OOM & OOM & OOM & OOM & OOM & OOM & OOM & NA\\
&~~~StaticPCA 0.8x & 14.21       & 27.54            & 17.41    & 19.77    & 5.11    & 0.25 & 5.65   & 3.01           & 0.5                    & 13.41 & 18.84       & 11.43 \\
&~~~Palu 0.6x & 17.17       & 23.63            & 9.71     & 10.22    & 5.3     & 0    & 3.86   & 3.16           & 7.1                    & 5.08  & 4.77        & 8.18 \\
&~~~\textbf{\kv{}}~0.6x  & 15.91       & 31.48            & 19.97    & 20.34    & 7.69    & 0    & 5.76   & 2.01           & 2.28                   & 16.06 & 21.15       & 12.97\\

&~~~\textit{OjaKV-PF}~0.6x   &  16.86       & 33.76            & 23.49    &  25.91   & 7.42   & 0  & 5.26   & 3.14            & 3.75                    & 14.65 & 16.79       & 13.73 \\

\specialrule{1pt}{2pt}{10pt}\specialrule{1pt}{2pt}{2pt}
\multirow{5}{*}{\rotatebox[origin=c]{90}{\fontsize{15}{100}\selectfont Llama-3.1-8B}}
&~~~Full KV   & 29.56 & 53.0 & 53.76 & 46.13 & 28.38 & 7.5 & 7.47 & 6.25 & 99.5 & 19.88 & 19.22 & \avgXI{29.56+53.0+53.76+46.13+28.38+7.5+7.47+6.25+99.5+19.88+19.22} \\
\cline{2-14}
&~~~Eigen-N 0.8x  & OOM & OOM & OOM & OOM & OOM & OOM & OOM & OOM & OOM & OOM & OOM & NA      \\
&~~~StaticPCA 0.8x   & 11.81       & 49.68            & 48.06    & 43.56    & 15.43   & 10   & 9.46   & 3              & 89.5                   & 21.36 & 22.46       & 29.48      \\
&~~~Palu 0.8x  & 30.54       & 32.91            & 21.38    & 18.07    & 14.76   & 71   & 4.07   & 3              & 93.14                  & 5.14  & 5.86        & 27.26 \\
&~~~\textbf{\kv{}}~0.8x & 12.34       & 48.89            & 49.26    & 44.69    & 17.47   & 9    & 9.59   & 3              & 89                     & 21.71 & 22.92       & 29.81     \\
&~~~\textit{OjaKV-PF}~0.8x   &   12.11      & 48.81            & 48.94    &  44.58   & 18.87   & 10  & 7.88   & 4            & 87.5                    & 21.2 & 22.35       &  29.66\\
\cline{2-14}
&~~~Eigen-N 0.6x   & OOM & OOM & OOM & OOM & OOM & OOM & OOM & OOM & OOM & OOM & OOM & NA      \\
&~~~StaticPCA 0.6x & 8.16        & 39.38            & 25.63    & 22.73    & 12.65   & 0.67 & 7.1    & 1.62           & 12.5                   & 18.76 & 23.48       & 15.70 \\
&~~~Palu 0.6x & 24.25       & 28.35            & 18.72    & 16.53    & 10.86   & 65   & 4.28   & 3              & 20.49                  & 2.81  & 3.23        & 17.96 \\
&~~~\textbf{\kv{}}~0.6x   & 9.73        & 43.94            & 35.95    & 34.49    & 15.21   & 3.5  & 7.78   & 3.5            & 33                     & 18.21 & 21.94       & 20.66 \\

&~~~\textit{OjaKV-PF}~0.6x   &    11.63     & 47.99            & 46.98    &  43.71   & 18.87   & 4  & 7.88   & 4            & 87.5                    & 21.59 & 22.97       &  28.46 \\

\specialrule{1pt}{2pt}{10pt}\specialrule{1pt}{2pt}{2pt}
\end{tabular}
}
\end{threeparttable}
\end{table*}

\subsection{RULER}
\paragraph{Setup.} We benchmark \kv{} on the RULER long-context retrieval suite using LongChat-7b-v1.5-32k. We evaluate performance on challenging input sequences of 16K tokens, creating significant GPU memory pressure. We report results under three cache budgets: uncompressed (Full), 20\% savings ($0.8\times$), and 40\% savings ($0.6\times$). For each compressed budget, we compare StaticPCA, Palu, and Eigen-N baselines against our context-aware \kv{}.

\paragraph{Results.} 
As shown in Table~\ref{tab:ruler_results}, the limitations of existing low-rank methods become apparent on the demanding RULER benchmark. The Eigen-N baseline is infeasible in this setting, as its incompatibility with FlashAttention leads to OOM errors at this sequence length. Palu fails entirely, producing zero accuracy across all tasks. StaticPCA, while functional, suffers substantial degradation—averaging only 28.37 at $0.8\times$ and 23.44 at $0.6\times$ compression.

In contrast, \kv{} achieves strong retrieval accuracy across all RULER subtasks. These results validate the effectiveness of our dynamic, context-adaptive framework.

\begin{table}[t]
\centering
\caption{Retrieval accuracy (\%) on the RULER using LongChat-7b-v1.5-32k.}
\vspace{-10pt}
\resizebox{\columnwidth}{!}{%
\scriptsize
\setlength{\tabcolsep}{5pt}
\renewcommand{\arraystretch}{1.2}
\label{tab:ruler_results}
\begin{tabular}{l|cccccccc}
\toprule
& \multicolumn{8}{c}{\textbf{prompt length 16K}} \\
\textbf{Method} & S1 & S2 & MK1 & MK2 & MQ & MV & QA & Avg \\
\midrule
FullKV & 100 & 99 & 91 & 74 & 70 & 71 & 13 & 74 \\
\midrule
Eigen\text{-}N 0.8x       & OOM  & OOM  & OOM  & OOM  & OOM & OOM & OOM & N/A \\
StaticPCA 0.8x & 68              & 17              & 23                & 4                 & 47.75            & 25.75            & 13.08 & 28.37            \\
Palu 0.8x & 0              & 0              & 0                & 0                 & 0            & 0         & 0 & 0   \\
\textbf{\kv{}} 0.8x    & 90              & 50              & 40                & 13                & 55.75            & 44.75            & 20.42 & 44.85  \\

\textit{OjaKV-PF} 0.8x    & 95              & 92              & 76                & 18                & 38            & 47            & 13.33 & 54.19  \\

\midrule
Eigen\text{-}N 0.6x       & OOM  & OOM  & OOM  & OOM  & OOM & OOM  & OOM & N/A \\
StaticPCA 0.6x   & 59              & 5               & 20                & 3                 & 30.75            & 31               & 15.33 & 23.44         \\
Palu 0.6x   & 0              & 0              & 0                & 0                 & 0            & 0         & 0 & 0  \\
\textbf{\kv{}} 0.6x    & 94              & 51              & 37                & 8                 & 32.25            & 34.25            & 23.42 & 39.99   \\
\textit{OjaKV-PF} 0.6x    & 97              & 91              & 71                & 15                & 31.75            & 36            & 13.33 & 50.7  \\
\bottomrule
\end{tabular}
}
\end{table}

\subsection{LongBench}

\label{subsec:longbench}

We further evaluate \kv{} on LongBench, a benchmark designed to test long-context inference across diverse tasks, including single and multi-document QA, few-shot learning, and code generation.\ The task input lengths vary from a few thousand to over 20K tokens.\ As shown in Table~\ref{tab:longbench}, \kv{} outperforms StaticPCA across both models and compression ratios on the majority of tasks.\ The performance advantage is less pronounced compared to RULER, potentially because LongBench tasks test for comprehension of a long, stable context. A case study demonstrating how OjaKV improves output quality is provided in Appendix~\ref{sec:case study}.

\subsection{Reasoning Task}

In this section we present the experiment setup and evaluate OjaKV using reasoning models. Specifically, we use the distilled reasoning models from R1-Distill, Deepseek-R1-Distill-Llama3-8b \citep{deepseekai2025deepseekr1incentivizingreasoningcapability} as the base models. We evaluate OjaKV and baseline methods on math and reasoning task AIME2025.

\begin{table}[htbp]
    \caption{Performance comparison of reasoning models across AIME2025 benchmark. Accuracy is reported in percent (\%).}
    \centering
    \scriptsize 
    \begin{tabular}{l|ccc}
    \toprule
    Model & Approach & AIME2025 \\
    \midrule
    \multirow{3}{*}{Llama} & Vanilla & 43.3 \\
\cmidrule{2-3}
& Eigen-N & 0 \\
& StaticPCA & 0 \\
& Palu &  0 \\
& \textbf{OjaKV} & \textbf{13.0} \\

\bottomrule
    \end{tabular}
    \label{table: aime}
\end{table}

Reasoning tasks present a unique challenge for KV cache compression: the input prompt is typically short (containing only the problem statement), while the model generates extremely long chains of thought during decoding. In this setting, all methods retain full-rank representations for the prompt tokens, but must compress the vast majority of tokens generated during decoding. This creates a stringent test for low-rank approximation quality, as the reasoning process continuously introduces new semantic patterns that a static basis cannot capture.

As shown in Table~\ref{table: aime}, static compression methods (Eigen-N, StaticPCA, and Palu) completely fail on AIME2025, achieving 0\% accuracy.\ In contrast, OjaKV maintains 13.0\% accuracy by continuously adapting its low-rank bases via online Oja updates throughout decoding. While there remains a gap to the vanilla baseline (43.3\%), OjaKV is the only compression method that preserves meaningful reasoning capability, demonstrating the critical importance of online subspace adaptation for long-form generation tasks.

\subsection{Efficiency}
\label{subsec:efficiency}

\begin{figure}[htbp]
    \centering
    \begin{subfigure}[b]{0.48\linewidth}
        \centering
        \includegraphics[width=\textwidth]{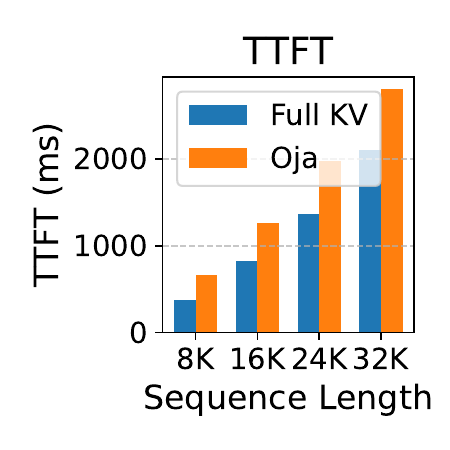}
        \label{fig:ttft_comparison}
    \end{subfigure}
    \hfill
    \begin{subfigure}[b]{0.48\linewidth}
        \centering
        \includegraphics[width=\textwidth]{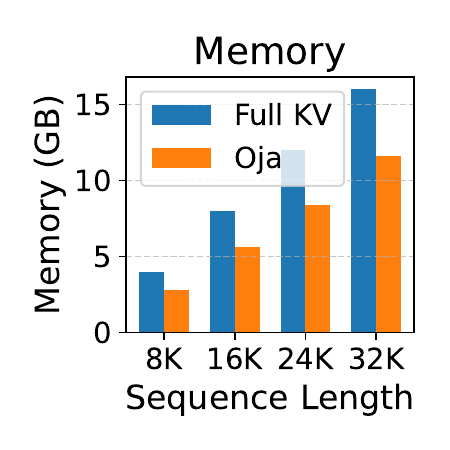}
        \label{fig:memory_comparison}
    \end{subfigure}
    \caption{Efficiency comparison of Full KV and OjaKV (60\% compression).\ }
    \label{fig:efficiency_comparison_oja}
\end{figure}

We compare OjaKV against the Full KV cache on Llama-3.1-8B-Instruct in terms of latency
and GPU memory (Figure~\ref{fig:efficiency_comparison_oja}).\ The Oja update introduces some overhead to TTFT.
At 32K tokens, TTFT increases from 2102\,ms (Full KV) to 2801\,ms (OjaKV).\ In contrast, memory
usage, which is the limiting factor for long-context inference, decreases from 16\,GB to 11.6\,GB
at 32K tokens.\ This memory reduction enables longer inputs under the same budget.

\begin{table}[htbp]
\centering
\scriptsize
\caption{Decoding latency comparison. $T$ denotes the Oja update interval (in tokens) during the decoding phase.}
\label{tab:latency}
\begin{tabular}{lc}
\toprule
\textbf{Method} & \textbf{Latency (ms/token)} \\
\midrule
Full KV & 36.48 \\
Static PCA & 36.60 \\
\midrule
OjaKV ($T=64$) & 41.38 \\
OjaKV ($T=128$) & 41.09 \\
OjaKV ($T=256$) & 40.44 \\
\bottomrule
\end{tabular}
\end{table}

Table~\ref{tab:latency} shows that OjaKV achieves online adaptation with modest latency overhead (10.9--13.4\%), 
while StaticPCA remains nearly as fast as FullKV. 
Notably, if one were to periodically recompute the PCA basis from scratch (as in StaticPCA) every $T$ tokens, 
the overhead would be substantially higher than OjaKV's incremental updates. 
The update interval $T$ thus provides a practical knob to balance basis freshness and computational efficiency.

\section{Ablation Studies}
To investigate the contribution of each component in OjaKV, we conduct a progressive ablation study on Llama-2-7B at 0.6x compression ratio. The results are summarized in Table~\ref{tab:ablation_ruler}. 


\begin{table}[htbp]
    \centering
    \caption{Ablation study on \textbf{RULER} benchmark. We report effective context length and accuracy (\%) on representative tasks.}
    \label{tab:ablation_ruler}
    \scriptsize 
    \setlength{\tabcolsep}{6pt}
    \begin{tabular}{l|cccc|c}
    \toprule
    \textbf{Configuration} & \textbf{S1} & \textbf{S2} & \textbf{MK1} & \textbf{MQ} & \textbf{Avg} \\
    \midrule
    StaticPCA (Baseline)     & 68 & 17 & 23 & 47.7 & 38.9\\
    + Hybrid Storage         & 88 & 47 & 39 & 54.5 & 57\\
    + Hybrid \& Online (\textbf{\kv{}}) & \textbf{90} & \textbf{50} & \textbf{40} & \textbf{55.8} & \textbf{60}\\
    \bottomrule
    \end{tabular}

\end{table}


Hybrid storage contributes the largest gain, improving average accuracy from 38.9\% to 57\%. This confirms that retaining high-error tokens at full rank is essential for preserving information poorly captured by the low-rank basis. Online Oja updates provide an additional improvement, demonstrating that adaptive subspace tracking offers complementary benefits.

%% file: Sections/Chapter_6.tex
\section{Conclusion}
In this work, we addressed the critical KV cache memory bottleneck in long-context LLMs. We introduced \textbf{\kv{}}, a novel framework that integrates a \textbf{hybrid storage policy} preserving critical tokens in full rank, and a lightweight, Oja-based \textbf{online update scheme} to adapt the low-rank subspace for remaining tokens.

Our extensive experiments show that \kv{} consistently outperforms strong static baselines at aggressive compression ratios. Crucially, our evaluation is one of the first to comprehensively assess low-rank methods on challenging, \textbf{generation-based} long-context tasks, revealing that while naive uniform compression can harm generation quality, \kv{}'s hybrid policy effectively mitigates this by strategically preserving key tokens. OjaKV demonstrates the largest gains on very challenging long-context benchmarks, confirming the value of online subspace adaptation. With full compatibility with FlashAttention and multiplicative savings when combined with token-eviction methods, \kv{} establishes a practical, parameter-free paradigm for efficient long-context LLM inference.

\paragraph{Future Work.}
A promising direction is to replace fixed update schedules with adaptive ones. In our current implementation, Oja updates during decoding are triggered every $T$ steps using a manually chosen interval. However, decode-stage reconstruction error provides a natural signal for when the current low-rank subspace is no longer well aligned with the evolving context. In preliminary analysis, we observe that this error often exhibits sparse, spike-like behavior rather than growing monotonically, suggesting that subspace updates may only be necessary at a small number of decoding positions. This motivates an error-triggered update rule that dynamically adjusts update frequency based on reconstruction error statistics, potentially improving the trade-off between adaptation quality and runtime overhead. More broadly, another promising direction is to replace fixed hyperparameters with dynamic schedules, such as adapting learning rates, update frequency, and buffer size based on activation shift or subspace drift.